\documentclass[12pt]{article}
\usepackage[a4paper,total={18cm,27cm}]{geometry}
\usepackage{graphicx}
\usepackage{authblk}
\usepackage{amsmath}
\usepackage{amsfonts} 
\usepackage[utf8]{inputenc}
\usepackage{hyperref}
\usepackage{verbatim}
\usepackage{color,soul}
\usepackage{caption}
\usepackage{xcolor}
\usepackage{framed}
\colorlet{shadecolor}{yellow!20}

\begin{document}

% ******************************************
% TITLE AND AUTHORS
% ******************************************

\title{
Classification at the Accuracy Limit -\\
Facing the Problem of Data Ambiguity
}

\author[1,2]{Claus Metzner}
\author[1,3]{Achim Schilling}
\author[4]{\\Maximilian Traxdorf}
\author[1]{Konstantin Tziridis}
\author[1]{Holger Schulze}
\author[1,3,5]{\\Patrick Krauss}

\affil[1]{\small Neuroscience Lab, University Hospital Erlangen, Germany}

\affil[2]{\small Biophysics Lab, Friedrich-Alexander University Erlangen-Nuremberg, Germany}

\affil[3]{\small Cognitive Computational Neuroscience Group, Friedrich-Alexander University Erlangen-Nuremberg, Germany}

\affil[4]{\small 
Department of Otorhinolaryngology, Paracelsus Medical University, Nuremberg, Germany}

\affil[5]{\small Pattern Recognition Lab, Friedrich-Alexander University Erlangen-Nuremberg, Germany}

\maketitle

% ******************************************
% ABSTRACT
% ******************************************

\begin{abstract}
\large
Data classification, the process of analyzing data and organizing it into categories, is a fundamental computing problem of natural and artificial information processing systems. Ideally, the performance of classifier models would be evaluated using unambiguous data sets, where the 'correct' assignment of category labels to the input data vectors is unequivocal. In real-world problems, however, a significant fraction of actually occurring data vectors will be located in a boundary zone between or outside of all categories, so that perfect classification cannot even in principle be achieved. We derive the theoretical limit for classification accuracy that arises from the overlap of data categories. By using a surrogate data generation model with adjustable statistical properties, we show that sufficiently powerful classifiers based on completely different principles, such as perceptrons and Bayesian models, all perform at this universal accuracy limit. Remarkably, the accuracy limit is not affected by applying non-linear transformations to the data, even if these transformations are non-reversible and drastically reduce the information content of the input data. We compare emerging data embeddings produced by supervised and unsupervised training, using MNIST and human EEG recordings during sleep. We find that categories are not only well separated in the final layers of classifiers trained with back-propagation, but to a smaller degree also after unsupervised dimensionality reduction. This suggests that human-defined categories, such as hand-written digits or sleep stages, can  indeed be considered as 'natural kinds'.
\end{abstract}

\vspace{1cm}
\noindent Correspondence to {\em claus.metzner@gmail.com}

% ******************************************
% INTRODUCTION
% ******************************************

\clearpage
\section{Introduction}

Data classification -- e.g. object recognition -- is a fundamental computing problem in machine learning and artificial intelligence. Large-scale classification competitions such as the annual ImageNet challenge \cite{krizhevsky2012imagenet, russakovsky2015imagenet}, where a super-human accuracy of 95\% has been achieved within about 5 years of steady progress, have contributed greatly to the general popularity of machine learning. Understandably, ImageNet is mostly discussed in the context of technical improvements regarding the classification methods which enabled this drastic boost of performance. But it also illustrates some fundamental problems that arise when computers are to create models of human-defined data categories:
For example, the fact that classification accuracies are typically leveling off at values below 100\% does not necessarily reflect a limitation of the algorithms, but instead may reveal the classification limits of the humans who provided the ground truth data. Indeed, in the case of ImageNet, the massive work of annotating millions of images had been crowd-sourced using Amazon Mechanical Turk, and so a large number of individuals were involved in the labeling process, individuals who may place certain ambiguous images into different categories. This problem of ambiguity due to non-rigorously defined object categories is most pronounced in biological and medical data, where sample-to-sample variations are notoriously large.  

In this work, we use artificially generated surrogate data, as well as real-world bio-medical data, to explore the implications of this inevitable data ambiguity. We demonstrate that the overlap of data classes leads to a theoretical upper limit of classification accuracy, a limit that can be mathematically computed in low-dimensional examples and which depends in a systematic way on the statistical properties of the data set. We find that sufficiently powerful classifier models of different kinds all perform at this same upper limit of accuracy, even if they are based on completely different operating principles. Interestingly, this accuracy limit is not affected by applying certain non-linear transformations to the data, even if these transformations are non-reversible and drastically reduce the information content (entropy) of the input data. 

In a next step, the same three models that reached the common classification limit for artificial data are now applied to human EEG data measured during sleep. In a pre-processing step, two kinds of features are extracted from raw EEG signals, yielding different marginal distributions and mutual correlations. It turns out that a more complex Bayesian model, based on correlated multi-variate Gaussian likelihoods (CMVG), performs worse than two other models (naive Bayes, perceptron), because the statistical properties of the pre-processed features do not match those of the likelihoods. In contrast, the perceptron and the naive Bayes model still show very similar classification accuracies, indicating that both reach the theoretical accuracy limit for sleep stage classification.

Finally, we address the question whether typical human-defined object categories can also be considered as 'natural kinds', that is, whether the data vectors in input space have a built-in cluster structure that can be detected by objective machine-learning models even in non-labeled data. For this purpose, we use as real-world examples the MNIST data set \cite{deng2012mnist}, as well as the above EEG sleep data. We find that a simple visualization by multi-dimensional scaling (MDS) \cite{torgerson1952multidimensional, kruskal1964nonmetric,kruskal1978multidimensional,cox2008multidimensional} already reveals an inherent cluster structure of the data in both cases. Interestingly, the degree of clustering, quantified by the general discrimination value (GDV) \cite{krauss2018statistical, schilling2021quantifying}, can be enhanced by a step-wise dimensionality reduction of the data, using an autoencoder that is trained in an unsupervised manner. A perceptron classifier with a layer design comparable to the autoencoder, trained on the same data in a supervised fashion, achieves as expected a much stronger cluster separation. However, the enhancement of clustering by unsupervised data compression, combined with automatic labeling methods, could be a promising way to automatically detect 'natural kinds' in non-labeled data. 

% ******************************************
% METHODS
% ******************************************

\clearpage
\section{Methods}

\subsection*{Part 1: Accuracy limit}

\subsubsection*{Derivation of theoretical accuracy limit}

Classification is the general problem of assigning a discrete class label $i=1\ldots K$ to each given input data $\vec{x}$, where the latter is considered as a vector with $N$ real-valued components $x_{n=1\ldots N}$. Such a discrimination is possible when the conditional probability distributions $p_{gen}(\vec{x}\;|\;i)$ of data vectors, here called {\bf 'generation densities'}, are different for each of the possible data classes $i$. In the simple case of a two- or three-dimensional data space, each data class can be visualized as a 'point cloud' (See Fig.\ref{figure_1}(a,b) for examples), and either the shapes or the center positions of these point clouds must vary sufficiently in order to facilitate a reliable classification. However, since the data generation process typically involves not only the system of interest (which might indeed have $K$ well-distinguished modes of operation), but also some measurement or data transmission equipment (which introduces noise into the data), a certain 'overlap' of the different data classes is usually not avoidable.

\vspace{0.2cm}\noindent A classifier is receiving the data vectors $\vec{x}$ as input and computes a set of $K$ {\bf 'classification probabilities'} $q_{cla}(j\;|\;\vec{x})$, quantifying the belief that $\vec{x}$ belongs to $j$. They are normalized to one over all possible classes, so that $\sum_{j=1}^K q_{cla}(j\;|\;\vec{x}) = 1\; \forall\; \vec{x}$.

\vspace{0.2cm}\noindent  We can now define a {\bf 'confusion density'} as the product
\begin{equation}
C_{ji}(\vec{x}) = q_{cla}(j\;|\;\vec{x})\;p_{gen}(\vec{x}\;|\;i).
\end{equation}
It can be interpreted as the probability density that the generator is producing data vector $\vec{x}$ under class $i$, which is then assigned to class $j$ by the classifier. Because there is usually a very small but non-zero probability density that {\em any} vector $\vec{x}$ can occur under {\em any} class $i$, we expect that the non-diagonal elements $C_{j\!\neq\!i}(\vec{x})$ are larger than zero as well. These non-diagonal confusion densities will have their largest values in regions of data space where the classes $i$ and $j$ overlap (See Fig.\ref{figure_1}(e,f) for examples).

\vspace{0.2cm}\noindent  By integrating the confusion density over all possible data vectors $\vec{x}$,
\begin{equation}
C_{ji} = \int C_{ji}(\vec{x})\; d\vec{x},
\label{cij}
\end{equation}
we obtain the {\bf 'confusion matrix'} of the classifier, which comes out properly normalized, so that $\sum_{j=1}^K C_{ji} = 1\; \forall\; i$. The confusion matrix $C_{ji}$ therefore is the probability that a data point originating from class $i$ is assigned to class $j$.

\vspace{0.2cm}\noindent   Assuming for simplicity that all data classes appear equally often, we can compute the {\bf accuracy} $A$ of the classifier as the average over all diagonal elements of the confusion matrix:
\begin{equation}
A = \frac{1}{K}\sum_{i=1}^K C_{ii}.
\end{equation}

\vspace{0.2cm}\noindent In the following, we are particularly interested in the {\bf theoretical limit of the classification accuracy}, denoted by $A_{max}$. We therefore consider an ideal classifier that has learned the exact generation densities $p_{gen}(\vec{x}\;|\;i)$. In this case, the {\bf 'ideal classification probability'} corresponds to the Bayesian posterior
\begin{equation}
q_{cla}(\;j\;|\;\vec{x}\;) = \frac{p_{gen}(\;\vec{x} \;|\; j\;)}{\sum_k p_{gen}(\;\vec{x} \;|\; k\;)}. 
\end{equation}

\vspace{0.2cm}\noindent In our numerical experiments, we will use classifiers that output a definite class label $j$ for each given data vector $\vec{x}$, corresponding to most probably class. To compute the theoretical accuracy maximum for such a model, we replace $q_{cla}$ by the binary {\bf 'class indicator function'}
\begin{equation}
\hat{q}_{cla}(\;j\;|\;\vec{x}\;) = \delta_{jk}\;\;\mbox{with}\;\;k = \mbox{argmax}_c\; q_{cla}(\;c\;|\;\vec{x}\;).
\end{equation}
It has the value $1$ for all data points $\vec{x}$ assigned to class $j$, and the value $0$ for all other data points (See Fig.\ref{figure_1}(c,d) for examples). When the ideal accuracy $A$ is evaluated using $\hat{q}_{cla}$ instead of $q_{cla}$, the result can be directly compared with numerical accuracies based on one-hot classifier outputs.

\subsubsection*{Numerical evaluation of $A_{max}$}

In Fig.\ref{figure_1}, the above quantities have been numerically evaluated for a simple Gaussian test data set. For this purpose, the two-dimensional integral \ref{cij} has been evaluated numerically on a regular grid of linear spacing 0.01, ranging from -8 to +8 in each feature dimension. 

%\vspace{0.2cm}\noindent It is often advantageous to feed classifiers not directly with high-dimensional raw data vectors $\vec{x}$, but with pre-processed vectors $\vec{u} = F(\vec{x})$ that have a much smaller number $D\ll N$ of components $u_{f=1\ldots D}$, in the following called {\bf features}. A suitable pre-processing can accelerate the training procedure of the classifier and can even increase the classification accuracy, provided the theoretical limit has not already been reached. Note that for suitable invertible mappings $\vec{u} = F(\vec{x})$, the pre-processing amounts to a variable transform of the integral Eq. \ref{cij}, which leaves the value of the integral unchanged. The theoretical limit of classification accuracy will therefore remain invariant under a large class of linear or nonlinear data transformations.

\subsubsection*{Classifiers and input data}

\vspace{0.2cm}\noindent In the following subsections, we provide the implementation details for the different classifier models that are compared in this work. The input data for these models is given as lists of $D$-dimensional feature vectors $\vec{u} = (u_1,u_2,\ldots,u_f,\ldots,u_D)$, each belonging to one of $K$ possible classes $c$. In the case of artificially generated data, these lists contain 10000 feature vectors distributed equally over the data classes. They are split randomly into training (80\%) and test (20\%) data sets.

\subsubsection*{Perceptron model}

The perceptron model is implemented using Keras/Tensorflow. It has one hidden layer, containing $N_{neu}=100$ neurons with RELU activation function. The output layer has $N_{out}$ neurons with softmax activation function, where $N_{out}=D$ corresponds to the number of data classes. The loss function is categorical crossentropy. We optimize the perceptron on each training data set using the Adams optimizer over at least 10 epochs with a batch size of 128 and a validation split of 0.2. After training, the accuracy of the perceptron is evaluated with the independent test data set.

\subsubsection*{Naive Bayesian model}

The naive Bayesian model is implemented using the Python libraries Numpy and Scipy.

\vspace{0.2cm}\noindent  In the training phase, the training data set is sorted according to the $K$ class labels $c$. Then an individual Gaussian Kernel Density (KDE) approximation (Scott method) is computed for each feature $f$ and class label $c$, corresponding to the empirical marginalized probability densities $p_{f,c}(u_f)$. 

\vspace{0.2cm}\noindent In the testing phase, the accuracy of the model is evaluated with the independent test data set as follows:  According to the naive Bayes approach, the global likelihood $L(\vec{u}\;|\;c)$ of a data vector $\vec{u}=(u_1,u_2,\ldots,u_D)$ under class $c$ is approximated by a product of the marginalized probabilities, so that
\begin{equation}
L(\vec{u}\;|\;c) = \prod_{f=1\ldots D} p_{f,c}(u_f).
\end{equation}
Since we assume a flat prior probability ($P_{prior}(c)=1/K$) over the data classes, the posterior probability of data class $c$, given the input data vector $\vec{u}$, is given by
\begin{equation}
P_{post}(c\;|\;\vec{u}) = \frac{L(\vec{u}\;|\;c)}{\sum_{i\!=\!1}^K\;L(\vec{u}\;|\;i)}
\end{equation}

\subsubsection*{Naive Bayesian model with Random Dimensionality Expansion (RDE)}

Since the naive Bayesian model takes into account only the marginal feature distributions $p_{f,c}(u_f)$, it cannot distinguish data classes which accidentally have identical $p_{f,c}(u_f)$ distributions, but differ in the correlations between the features. In principle, this problem can be fixed by multiplying the $D$-dimensional input vectors $\vec{u}$ by a random $D_2 \times D$ matrix $\textbf{M}$, for example with normally distributed entries $M_{ij}\propto N(\mu=0,\sigma=1)$, which yields transformed vectors $\vec{v} = \textbf{M} \vec{u}$. Provided that $D_2\gg D$, at least some of the new feature linear combinations $v_f$ will have marginal distributions that vary between the data classes.  

\subsubsection*{CMVG Bayesian model}

The Correlated Multi-Variate Gaussian (CMVG) Bayesian model is also implemented using the Python libraries Numpy and Scipy. 

\vspace{0.2cm}\noindent  In the training phase, the training data set is sorted according to the two class labels $c$. Then, for each class label $c$, we compute the mean values $\mu_f^{(c)}$ of the features $f=1\ldots D$, as well as the covariances $\Sigma_{fg}^{(c)}$ between features $f$ and $g$. These quantities are packed as one vector ${\bf \mu}^{(c)}$ and one matrix ${\bf \Sigma}^{(c)}$ for each class $c$.

\vspace{0.2cm}\noindent  In the testing phase, the global likelihood $L(\vec{u}\;|\;c)$ of a data vector $\vec{u}=(u_1,u_2,\ldots,u_D)$ under class $c$ is computed as the correlated, multi-variate Gaussian probability density

\begin{equation}
L(\vec{u}\;|\;c) = p_{cmvg}\left( \vec{u} \;,\; {\bf\mu}\!=\!{\bf\mu}^{(c)} \;,\;
{\bf \Sigma}\!=\!{\bf \Sigma}^{(c)}
\right).
\end{equation}
Since we assume a flat prior probability ($P_{prior}(c)=1/2$) for the two data classes, the posterior probability of data class $c$, given the input data vector $\vec{u}$, is given by
\begin{equation}
P_{post}(c\;|\;\vec{u}) = \frac{L(\vec{u}\;|\;c)}{\sum_{i\!=\!1}^K\;L(\vec{u}\;|\;i)}
\end{equation}

\subsection*{Part 2: The DSC data model}

We consider an artificial classification problem with two multivariate Gaussian data classes $c\in\left\{0,1\right\}$ and with statistical properties that can be tuned by {\bf three control quantities}: the {\bf dimensionality} $D$ of the feature space, the {\bf separation} $S$ between the centers of the point clouds, and the {\bf correlation} $C$ between features (within the same class), which is  associated with the shape of the point cloud. The generation of artificial data within this DSC model works as follows:

\vspace{0.2cm}\noindent Starting from a given triple $D,S,C$ of control quantities, we first generate $N_{rep}$ independent {\bf parameter sets} $[\mu_f^{(c)}, \Sigma_{fg}^{(c)}]$ that describe the statistical properties of the two classes $c\in\left\{0,1\right\}$. Here, $\mu_f^{(c)}$ is the mean value of feature $f$ in class $c$, and $\Sigma_{fg}^{(c)}$ is the covariance of features $f$ and $g$ in class $c$.

\vspace{0.2cm}\noindent The mean values $\mu_f^{(c=0)}$ in class $0$ are always set to zero, whereas the mean values $\mu_f^{(c=1)}$ in class $1$ are random numbers, drawn from a uniform distribution with values in the range from 0 to $S$. The separation quantity $S$ is therefore the maximum distance between corresponding feature mean values in each dimension $f$.

\vspace{0.2cm}\noindent The diagonal elements $\Sigma_{ff}^{(c)}$ of the symmetric covariance matrix are set to 1 in both classes. The off-diagonal elements $\Sigma_{f\neq g}$ are assigned independent, continuous random numbers $x$, drawn from a box-shaped probability density distribution $q(x,C)$ that depends on the correlation quantity $C$ as follows:

\begin{eqnarray}
 q(x,C) &=& \mbox{uniform}[0\;,\;C]\;\;\;\;\;\;\mbox{for}\;C\le1 \nonumber\\
 q(x,C) &=& \mbox{uniform}[C\!-\!1\;,\;1]\;\mbox{for}\;C>1
\end{eqnarray}

\vspace{0.2cm}\noindent For $C=0$, the distribution $q(x,C)$ peaks at $x=0$, so that $\Sigma_{ij}$ becomes a diagonal unit matrix. For $C=1$, the distribution $q(x,C)$ is uniform in the range $[0,1]$, and for $C=2$ it peaks at $x=1$, leading to $\Sigma_{ij}=1$. A plot of the distribution is shown in \ref{figure_2}(b).

\vspace{0.2cm}\noindent According to the parameter set $[\mu_f^{(c)},\Sigma_{fg}^{(c)}]$, we then generate for each of the two classes $c$ a number $N_{vec}/2$ of random, Gaussian data vectors $\vec{u}(t)$, in which the $D$ components (features) are correlated to a degree controlled by quantity $C$. In the limiting case $C=0$, the $D$ time series $u_{f}(t)$ become statistically independent, whereas for $C=1$, the time series become fully correlated and thus identical. The total number of $N_{vec}$ data vectors is combined to a complete data set, in which vectors from the two classes (with corresponding labels $c$) appear in random order. By this way, we obtain for each triple $D,S,C$ of control quantities a total number of $N_{rep}$ independent data sets, each consisting of $N_{vec}$ data vectors. Since each data set obeys its own random parameters $[\mu_f^{(c)},\Sigma_{fg}^{(c)}]$, the DSC model reflects some of the heterogeneity of typical real world data. Finally, we split each data set into a training set (80\%) and a test set (20\%).

\vspace{0.2cm}\noindent Before applying different types of classifiers to the DSC data sets, we test that the feature correlations and the class separation can be controlled reliably and over a sufficiently large range, using the quantities $C$ and $S$ (Fig.\ref{figure_2}(d)).

\subsubsection*{Control of feature correlations by quantity $C$}

To evaluate correlation control, we fix the quantities $D=10$ and $S=1.0$ (Note that the separation has no effect on the correlations)  and vary $C$ over the complete available range of supported values from 0 to 2. For each $C$, we generate $N_{rep}=100$ independent data sets, each consisting of $N_{vec}=10000$ data vectors. For each data set, we estimate the empirical covariance matrix $\Sigma_{ij}^{(0)}$ of class 0. Because the matrix is symmetric, we compute the root-mean-square (RMS) average of all matrix elements above the diagonal. The blue line in (Fig.\ref{figure_2}(d)) shows for each $C$ the mean RMS, averaged over the $N_{rep}=100$ repetitions (The latter are shown as gray dots). We find an almost linear relation between $C$ and the mean RMS. In particular, we can realize the full range of correlations, including the limiting cases of independently fluctuating features (for $C=0$) and identically fluctuating features (for $C=2$).

\subsubsection*{Control of class separation by quantity $S$}

To evaluate separation control, we fix the quantities $D=10$ and $C=0.5$ and vary $S$ between 0 and 10. For each $S$, we generate $N_{rep}=100$ independent data sets, each consisting of $N_{vec}=10000$ data vectors. For each (labeled) data set, we compute the general discrimination value (GDV), a quantity that has been specifically designed to quantify the separation between classes in high dimensional data sets \cite{krauss2018statistical, schilling2021quantifying}. The orange line in Fig.\ref{figure_2}(d) is the mean negative GDV, averaged over the $N_{rep}=100$ repetitions (The latter are shown as gray dots). 

\vspace{0.2cm}\noindent The GDV is computed as follows: We consider $N$ points $\mathbf{x_{n=1..N}}=(x_{n,1},\cdots,x_{n,D})$, distributed within $D$-dimensional space. A label $l_n$ assigns each point to one of $L$ distinct classes $C_{l=1..L}$. In order to become invariant against scaling and translation, each dimension is separately z-scored and, for later convenience, multiplied with $\frac{1}{2}$:
\begin{align}
s_{n,d}=\frac{1}{2}\cdot\frac{x_{n,d}-\mu_d}{\sigma_d}.
\end{align}
Here, $\mu_d=\frac{1}{N}\sum_{n=1}^{N}x_{n,d}\;$ denotes the mean, and $\sigma_d=\sqrt{\frac{1}{N}\sum_{n=1}^{N}(x_{n,d}-\mu_d)^2}$ the standard deviation of dimension $d$.
Based on the re-scaled data points $\mathbf{s_n}=(s_{n,1},\cdots,s_{n,D})$, we calculate the {\em mean intra-class distances} for each class $C_l$ 
\begin{align}
\bar{d}(C_l)=\frac{2}{N_l (N_l\!-\!1)}\sum_{i=1}^{N_l-1}\sum_{j=i+1}^{N_l}{d(\textbf{s}_{i}^{(l)},\textbf{s}_{j}^{(l)})},
\end{align}
and the {\em mean inter-class distances} for each pair of classes $C_l$ and $C_m$
\begin{align}
\bar{d}(C_l,C_m)=\frac{1}{N_l  N_m}\sum_{i=1}^{N_l}\sum_{j=1}^{N_m}{d(\textbf{s}_{i}^{(l)},\textbf{s}_{j}^{(m)})}.
\end{align}
Here, $N_k$ is the number of points in class $k$, and $\textbf{s}_{i}^{(k)}$ is the $i^{th}$ point of class $k$.
The quantity $d(\textbf{a},\textbf{b})$ is the euclidean distance between $\textbf{a}$ and $\textbf{b}$. Finally, the Generalized Discrimination Value (GDV) is calculated from the mean intra-class and inter-class distances  as follows:
\begin{align}
\mbox{GDV}=\frac{1}{\sqrt{D}}\left[\frac{1}{L}\sum_{l=1}^L{\bar{d}(C_l)}\;-\;\frac{2}{L(L\!-\!1)}\sum_{l=1}^{L-1}\sum_{m=l+1}^{L}\bar{d}(C_l,C_m)\right]
 \label{GDVEq}
\end{align}

\noindent whereas the factor $\frac{1}{\sqrt{D}}$ is introduced for dimensionality invariance of the GDV with $D$ as the number of dimensions.
In the case of two Gaussian distributed point clusters, the resulting discrimination value becomes $-1.0$ if the clusters are located such that the mean inter cluster distance is two times the standard deviation of the clusters.

\subsection*{Part 3: Comparing classifiers}

In Fig.\ref{figure_3}, we determine the average accuracy of the three classifier types (See part 1 of the Methods section) for different combinations of the DSC control parameters. For each parameter combination, 100 data sets are sampled from the superstatistical distribution. For every data set, consisting of 8000 training vectors and 2000 test vectors, the three classifiers are trained from the scratch and then evaluated. This results in 100 accuracies for each classifier and each parameter combination. We then compute the mean value of these 100 accuracies, and this is the average accuracy plotted as colored lines in Fig.\ref{figure_3}(c-f). The individual, non-averaged accuracies are plotted as gray points.

\subsection*{Part 4: Feature transformations}

In Fig.\ref{figure_4}, we return to a much simpler test data set, consisting of two 'spherical' Gaussian data clusters in a two-dimensional feature space, which are centered at $\vec{x}=(-\frac{1}{2},0)$ and $\vec{x}=(+\frac{1}{2},0)$. respectively. All three classifier types reach the theoretical accuracy limit of about 0.69 in this case.

\vspace{0.2cm}\noindent In this part we explore how certain non-linear transformations of the original features (that is, $(x_1,x_2) \longrightarrow (f(x_1),f(x_2))$) affect the classification accuracy. In particular, we investigate the cases $f(x)=\sin(x)$, $f(x)=\cos(x)$ and $f(x)=\mbox{sgn}(x)$. The signum function yields -1 for negative arguments and +1 for positive arguments. For the special case $x=0$ it would return zero, but this practically does not happen, as the features $x$ are continually distributed random variables.  

\subsection*{Part 5: Sleep EEG data}

For a real-world evaluation of classifier performance, we are using 68 multi-channel EEG data sets from our sleep laboratory, each corresponding to a full-night recording of brain signals from a different human subject. The data were recorded with a sampling rate of 256 Hz, using three separate channels F4-M1, C4-M1, O2-M1. In this work, however, the signals from these channels are pooled, effectively treating them as data sets of their own.

\vspace{0.2cm}\noindent  The participants of the study included 46 males and 22 females, with an age range between 21 and 80 years. Exclusion criteria were a positive history of misuse of sedatives, alcohol or addictive drugs, as well as untreated sleep disorders. The study was conducted in the Department of Otorhinolaryngology, Head Neck Surgery, of the Friedrich-Alexander University Erlangen-Nürnberg (FAU), following approval by the local Ethics Committee (323–16 Bc). Written informed consent was obtained from the participants before the cardiorespiratory polysomnography (PSG). 

\vspace{0.2cm}\noindent  After recording, the raw EEG data were analyzed by a sleep specialist accredited by the German Sleep Society (DGSM), who removed typical artifacts
\cite{tatum2011artifact} from the data and visually identified the sleep stages in subsequent 30-second epochs, according to the AASM criteria (Version 2.1, 2014) \cite{iber2007aasm,american2012aasm}. The resulting, labeled raw data were then used as a ground truth for testing the accuracy of the different classifier types.

\vspace{0.2cm}\noindent  In this work, we are primarily testing the ability of the classifiers to assign the correct sleep label $s$
(Wake, REM, N1, N2, N3) independently to each epoch, without providing further context information. Such a single-channel epoch consists of $30\times256=7680$ subsequent raw EEG amplitudes $x_{d,e}(t_n)$, where $d$ is the data set, $e$ the number of the epoch within the data set, and $t_n$ the $n$th recording time within the epoch. 

\vspace{0.2cm}\noindent In order to facilitate classification of these 7680-dimensional input vectors $\vec{x}_{d,e}$ by a simple Bayesian model, or by a flat two-layer perceptron with relatively few neurons, the vectors have to be suitably pre-processed and compressed down to feature vectors $\vec{u}_{d,e}$ of much smaller dimensionality $D\ll 7680$. 
\vspace{0.2cm}\noindent Instead of relying on self-organized (and thus 'black-box') features, we are using mathematically well-defined features with a simple interpretation. In particular, we are interested in the case where all $D$ components $u_f$ of a feature vector $\vec{u}$ are fundamentally of the same kind and only differ by some tunable parameter. 

\subsubsection*{Fourier features}

Our first type of feature estimates the momentary Fourier component of the raw EEG signal $x_{d,e}(t_n)$ at a certain, tunable frequency $\nu_f$: 

\begin{equation}
u_f = 
\sqrt{\left(\;
\sum_{n=1}^{7680} x_{d,e}(t_n)\cdot \cos(2\pi \nu_f t_n)
\right)^2 + 
\left(
\sum_{n=1}^{7680} x_{d,e}(t_n)\cdot \sin(2\pi \nu_f t_n)
\right)^2}.    
\end{equation}

\vspace{0.2cm}\noindent The set of frequencies $\nu_{f=1}\ldots\nu_{f=D}$ is in our case chosen as an equidistant grid between 0 Hz and 30 Hz, because our EEG system is filtering out the higher-frequency components of the raw signals above about 30 Hz. 

\subsubsection*{Correlation features}

Our second type of feature is the normalized auto-correlation coefficient of the raw EEG signal $x_{d,e}(t_n)$ at a certain, tunable lag-time $\Delta t_f$: 

\begin{equation}
u_f = \frac{\left\langle \left( x_{d,e}(t_n) - \overline{x}_{d,e}  \right)\cdot\left( x_{d,e}(t_n\!+\!\Delta t_f) - \overline{x}_{d,e} \right)  \right\rangle_n}{\sigma_{d,e}^2}.
\end{equation}

\vspace{0.2cm}\noindent Here, $ \overline{x}_{d,e}$ is the mean and $\sigma_{d,e}$ the standard deviation of the raw EEG signal within the epoch. The symbol $\left\langle \right\rangle_n$ stands for averaging over all time steps within the epoch. The set of lag-times $\Delta t_{f=1} \ldots \Delta t_{f=D}$ must be integer multiples of the recording time interval $\delta t = 1/256$ sec.

\subsection*{Part 6: Sleep stage detection}

In Fig.\ref{figure_6}, we investigate the performance of the three classifier types described in part 1 in the real-world scenario of personalized sleep-stage detection. For this purpose, the classifiers are trained and tested individually on each of our 68 full-night sleep recordings, using as inputs the same 6-dimensional Fourier- or correlation features as in Fig.\ref{figure_5} (Note that the aggregated distribution functions and covariance matrices in Fig.\ref{figure_5} have been computed by pooling over all data sets and therefore show a much more regular behavior than the individual ones).

\vspace{0.2cm}\noindent As a result, we obtain 68 accuracies for each combination of classifier type (Fig.\ref{figure_6}, rows) and used input feature (Fig.\ref{figure_6}, columns). The distributions of these accuracies are presented as histograms in the figure.

\subsection*{Part 7: Natural data clustering}

In Fig.\ref{figure_7}, we address the question whether typical real-world data sets have a built-in clustering structure that can be detected (and possibly enhanced) by unsupervised methods of data analysis. For this purpose, we visualize the clustering structure.
A frequently used method to generate low-dimensional embeddings of high-dimensional data is t-distributed stochastic neighbor embedding (t-SNE) \cite{van2008visualizing}. However, in t-SNE the resulting low-dimensional projections can be highly dependent on the detailed parameter settings \cite{wattenberg2016use}, sensitive to noise, and may not preserve, but rather often scramble the global structure in data \cite{vallejos2019exploring, moon2019visualizing}.
In contrats, multi-Dimensional-Scaling (MDS) \cite{torgerson1952multidimensional, kruskal1964nonmetric,kruskal1978multidimensional,cox2008multidimensional} is an efficient embedding technique to visualize high-dimensional point clouds by projecting them onto a 2-dimensional plane. Furthermore, MDS has the decisive advantage that it is parameter-free and all mutual distances of the points are preserved, thereby conserving both the global and local structure of the underlying data. 
When interpreting patterns as points in high-dimensional space and dissimilarities between patterns as distances between corresponding points, MDS is an elegant method to visualize high-dimensional data. By color-coding each projected data point of a data set according to its label, the representation of the data can be visualized as a set of point clusters. For instance, MDS has already been applied to visualize for instance word class distributions of different linguistic corpora \cite{schilling2021analysis}, hidden layer representations (embeddings) of artificial neural networks \cite{schilling2021quantifying,krauss2021analysis}, structure and dynamics of recurrent neural networks \cite{krauss2019analysis, krauss2019recurrence, krauss2019weight}, or brain activity patterns assessed during e.g. pure tone or speech perception \cite{krauss2018statistical,schilling2021analysis}, or even during sleep \cite{krauss2018analysis,traxdorf2019microstructure}. 
In all these cases the apparent compactness and mutual overlap of the point clusters permits a qualitative assessment of how well the different classes separate.

In addition, we measure the degree of clustering objectively by calculating the general discrimination value (GDV) \cite{krauss2018statistical,schilling2021quantifying}, described in part 2.

\vspace{0.2cm}\noindent For the clustering analysis we analyze two examples of 'natural data': One is the MNIST data \cite{deng2012mnist} set with 10 classes of handwritten digits, in which the input vectors are 784-dimensional (28x28 pixels) and have continuous positive values (between 0 and 1 after normalization). 

\vspace{0.2cm}\noindent As the second example we use, again, our full-night EEG recordings with the 5 data classes corresponding to the sleep stages Wake, REM, N1, N2, and N3. In order to reduce setup-differences between measurements, we first perform a z-transform over each individual full-night EEG recording, so that the one-channel EEG signal of each participant has now zero mean and unit variance. Next, in order to make the EEG data more comparable with MNIST, we produce one 784-dimensional input vector from each 30-second epoch of the EEG recordings in the following way: The 7680 subsequent one-channel EEG signals of the epoch are first transformed to the frequency domain using Fast Fourier Transform (FFT), yielding 3840 complex amplitudes. Since the phases of the amplitudes change in a highly irregular way between epochs, we discard this information by computing (the square roots of) the magnitudes of the amplitudes. We keep only the first 784 values of the resulting real-valued frequency spectrum, corresponding to the lowest frequencies. By pooling over all epochs and participants, we obtain a long list of these 784-dimensional input vectors. They are  globally normalized, so that the components in the list range between 0 and 1, just as in the MNIST case. Finally, the list is randomly split into train (fraction 0.8) and test (fraction 0.2) data sets. 

\vspace{0.2cm}\noindent It is possible to directly compute the MDS projection of the uncompressed 784-dimensional test data vectors into two dimensions, and also to calculate the corresponding GDV value that quantifies the degree of class separation (using the known sleep stage labeling). In Fig.\ref{figure_7}, these uncompressed data distributions are always shown in the left upper scatter plot of each two-by-two block.

\vspace{0.2cm}\noindent In this context, we also test if step-wise dimensionality reduction in an autoencoder leads to an enhanced clustering. The used autoencoder has RELU activation functions and 7 fully connected layers with the following numbers of neurons: 784,128,64,16,64,128,728. The mean squared error between input vectors and reconstructed vectors is minimized using the Adams optimizer. We also compute the MDS projections and GDV values for layers 2, 3 and 4 (the 16-dimensional bottleneck) of the autoencoder. In Fig.\ref{figure_7}, these three compressed data distributions are shown within the two-by-two blocks of scatter plots.

\vspace{0.2cm}\noindent As a reference for the resulting MDS projections and GDV values in the unsupervised autoencoder, we also process the two kinds of natural data with a perceptron that is trained in a supervised manner, so that it separates the known classes as far as possible. To make the perceptron comparable to the autoencoder, the first 4 layers (from the input to the bottleneck) are identical: Fully connected, RELU activations, and layer sizes 784,128,64,16. However, the decoder-part of the autoencoder is replaced by a softmax layer in the perceptron, which has either 10 (MNIST) or 5 (sleep) neurons. The perceptron is trained by back-propagation to minimize categorical cross-entropy between the true and predicted labels, using the Adams optimizer. Just as in the autoencoder, we compute MDS projections and GDV values for the first 4 perceptron layers.

% ******************************************
% RESULTS
% ******************************************

\clearpage
\section{Results}

\subsection*{Part 1: Accuracy limit}

\begin{figure}[ht!]
\centering
\includegraphics[width=12cm]{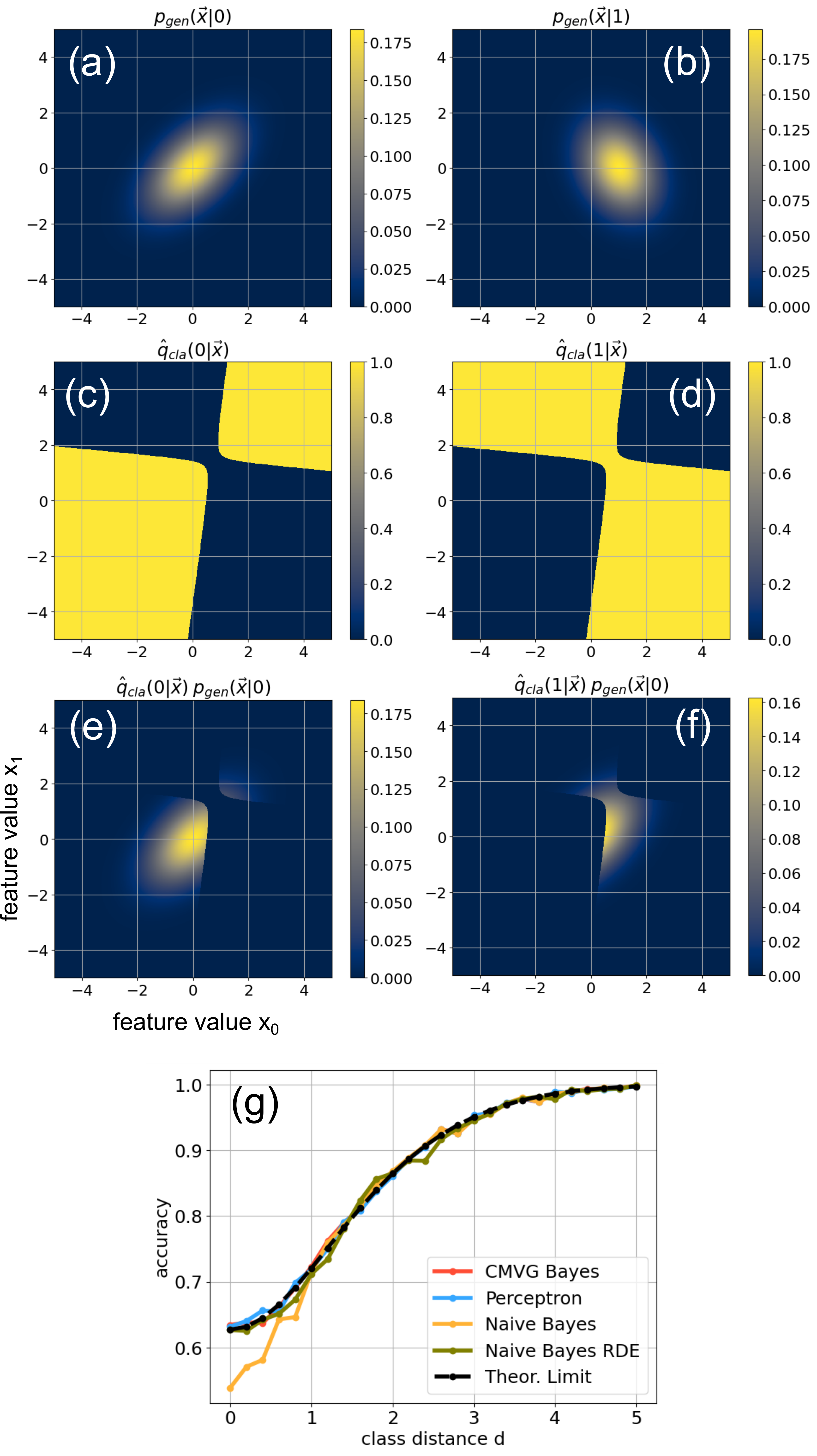}
\caption{Theoretical limit for classification accuracy. We consider two classes $i\in\{0,1\}$ in a two-dimensional feature space $\vec{x}$. The first row shows the probability densities $p_{gen}(\vec{x}|i)$ that data point $\vec{x}$ is generated under class $i=0$ (a) or class $i=1$ (b). The second row shows the binary probability distributions $\hat{q}_{cla}(j|\vec{x}) \in\{0,1\}$ that data point $\vec{x}$ is assigned to class $j=0$ (c) or class $j=1$ (d), assuming a perfect classifier. The third row shows the 'confusion densities' $\hat{q}_{cla}(j|\vec{x}) \; p_{gen}(\vec{x}|i)$ that data point $\vec{x}$ is generated under class $i=0$ and assigned to class $j=0$ (e) or class $j=1$ (f). Integrating this density over $\vec{x}$ yields the confusion matrix $C_{ji}$, from which the classification accuracy $A$ can be computed. Panel (g) in the fourth row shows the maximum possible classification accuracy (black curve) when the distance $d$ between the centers of the two data classes (a) and (b) in feature space is increased from 0 to 5. All classifier models considered in this paper (colored curves), except the Naive Bayes model (orange) at small distances, reach the theoretical accuracy limit.
\label{figure_1}}
\end{figure}

In order to demonstrate the existence of an accuracy limit in classification tasks, we assume a statistical process is generating data vectors $\vec{x}$ which are distributed in the input space (subsequently also called feature space) according to given {\bf generation densities} $p_{gen}(\vec{x}\;|\;i)$ that depend on the class $i$. For reasons of mathematical tractability and visual clarity, we start with a simple problem of two Gaussian data classes in a two-dimensional feature-space. We assume that class $i=0$ is centered at $\vec{x}=(x_1,x_2)=(0,0)$, whereas class $i=1$ is centered a distance $d$ away, at $\vec{x}=(d,0)$. As another discriminating property, the two class-dependent distributions are assumed to have different correlations between the features $x_1$ and $x_2$ (Compare Fig.\ref{figure_1}(a,b)). 

\vspace{0.2cm}\noindent As derived in the Methods section, an ideal classifier would divide the feature-space $\left\{\vec{x}\right\}$ among the two classes in a way that is perfectly consistent with the true generation densities $p_{gen}(\vec{x}\;|\;i)$. The resulting ideal assignment of a discrete class $j=0$ or $j=1$ to each data vector $\vec{x}$ can be described by binary {\bf class indicator functions} $\hat{q}_{cla}(\;j\;|\;\vec{x}\;)$ (Compare Fig.\ref{figure_1}(c,d)). 

\vspace{0.2cm}\noindent The latter two quantities can be combined to the {\bf confusion densities} $\hat{q}_{cla}(j|\vec{x}) \; p_{gen}(\vec{x}|i)$, which give the probability density that data point $\vec{x}$ is generated in class $i$ but assigned to class $j$ by the ideal classifier (Compare Fig.\ref{figure_1}(e,f)). The parts of feature space where the confusion density is large for $i\neq j$ correspond to the overlap regions of the data classes, and it is this overlap that makes the theoretical limit of the classification accuracy smaller than one.

\vspace{0.2cm}\noindent It is possible to compute the {\bf confusion matrix} of the ideal classifier by integrating the confusion densities over the entire feature space, which is feasible only in very low-dimension spaces. The confusion matrix, in turn, yields the {\bf theoretical accuracy limit} $A_{max}$ of the ideal classifier.
In our simple example, $A_{max}$ is expected to increase with the distance $d$ between the two data classes, as this separation reduces the class overlap. By numerically computing the integral over the two-dimensional feature space of our Gaussian test example, we indeed find a monotonous increase of $A_{max}=A_{max}(d)$ from about 0.62 at $d=0$ to nearly one at $d=5$ (Compare Fig.\ref{figure_1}(g, black line)).

\vspace{0.2cm}\noindent Our next goal is to apply different types of classifier models to data drawn from the generation densities $p_{gen}(\vec{x}\;|\;i)$ of the Gaussian test example above.

\vspace{0.2cm}\noindent As an example for a 'black box' classifier, we consider a {\bf perceptron} with one hidden layer (See Methods section for details). In the training phase, the connection weights of this neural network are optimized using the back-propagation algorithm. 

\vspace{0.2cm}\noindent As an example of a mathematically transparent, but simple classifier type, we consider a {\bf Naive Bayesian} model. Here, correlations between the input features are neglected, and so the global likelihood $L(\vec{u}\;|\;c)$ of a data vector $\vec{u}$, given the data class $c$, is approximated as the product of the marginal likelihood factors for each individual feature $f$ (See Methods section for details). In the 'training phase', the naive Bayesian classifier is simply estimating the distribution functions of these marginal likelihood factors, using Kernel Density Approximation (KDE). 

\vspace{0.2cm}\noindent Finally, we consider a {\bf Correlated Multi-Variate Gaussian (CMVG) Bayesian} model as an example of a mathematically transparent classifier that can also account for correlations in the data, but which assumes that all features are normally distributed (See Methods section for details). In the training phase, the CMVG Bayesian classifier has to estimate the mean values and covariances of the data vectors.

\vspace{0.2cm}\noindent When applying these three classifiers to the Gaussian test data, we indeed find that {\bf all models reach the same theoretical classification limit, even though their operating principles are very different} (Compare Fig.\ref{figure_1}(g)). The only exception is the Naive Bayes classifier at small class distances $d$ (Compare Fig.\ref{figure_1}(g, orange line)). This model fails because it can only use the marginal feature distributions, which happen to be identical for both classes in the case $d=0$. However, the problem can be easily fixed by multiplying the original two-dimensional feature vectors with a random, non-quadratic matrix (See Methods section for details) and thereby creating many new linear feature combinations, some of which usually have significantly different marginal distributions. Such a {\bf Random Dimensionality Expansion (RDE)}, as proposed in Yang et al. \cite{yang2021neural}, allows even the Naive Bayes model to reach the accuracy limit in strongly overlapping data classes (Compare Fig.\ref{figure_1}(g, olive line)).

\subsection*{Part 2: The DSC data model}

\begin{figure}[ht!]
\centering
\includegraphics[width=13cm]{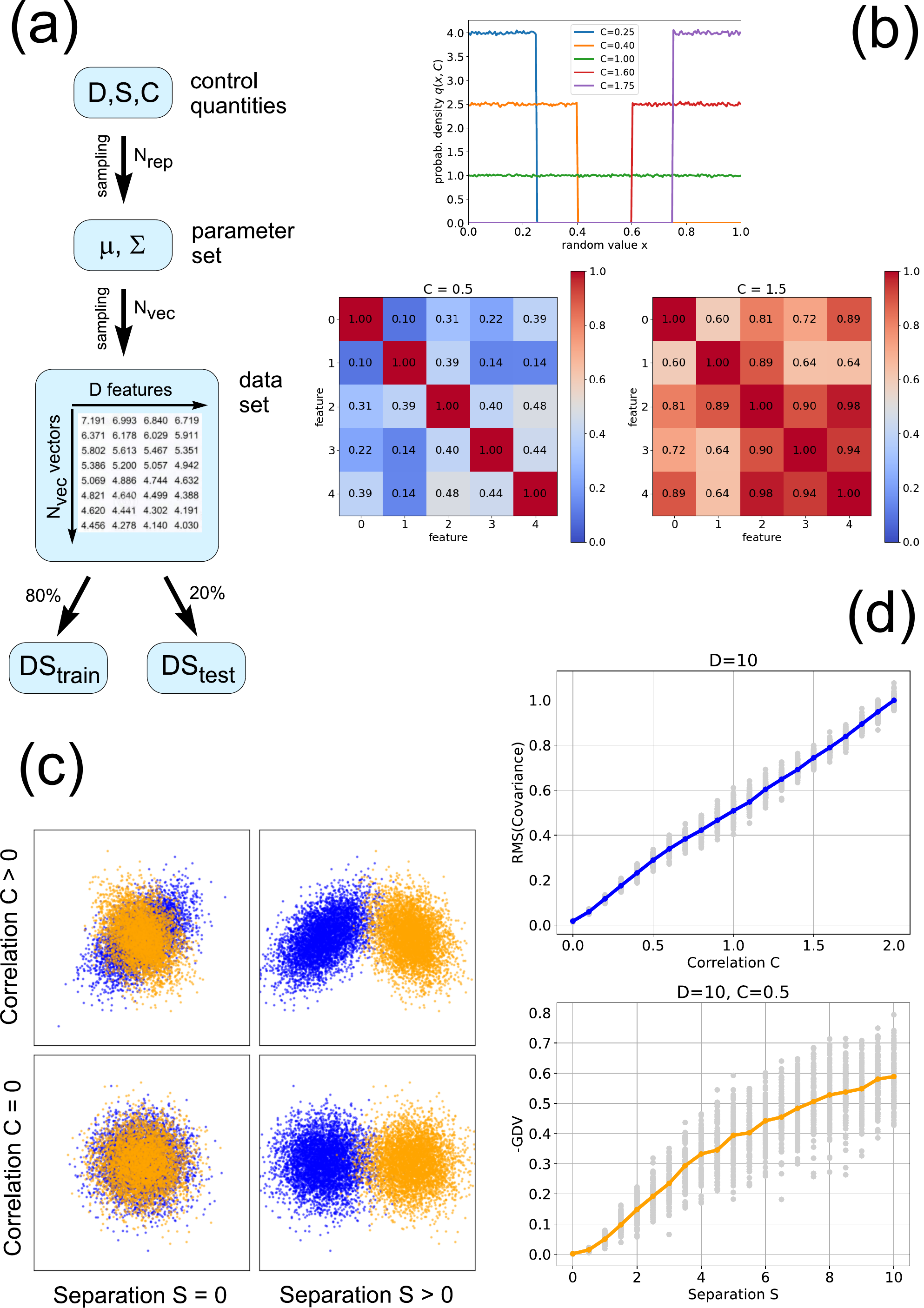}
\caption{The superstatistical DSC model for generating artificial data with dimensionality D, a prescribed separation S between the classes, and feature correlations C within the classes. 
{\bf (a)} Diagram depicting the hierarchical two-level structure of the model. For each triple of control quantities, $N_{rep}$ random parameter sets are generated. Then a data set of $N_{vec}$ vectors in $D$ dimensions is sampled according to each parameter set. 
{\bf (b)} The inter-feature correlations in each data set are described by covariance matrices $\Sigma_{ij}$, in which the level of correlations is controlled by the quantity $C\in[0,2]$. The off-diagonal elements $\Sigma_{i\neq j}$ are drawn from the $C$-dependent distribution $q(x,C)$ shown in the top. Setting, for example, $C=0.5$, these elements range from 0 to 0.5 (lower left), and for $C=1.5$ they range from 0.5 to 1 (lower right). 
{\bf (c)} Visualization of the two data classes for a case with D=2 dimensions. The classes correspond to point clouds in feature space, where quantity S affects the distance between the means and quantity C affects the shape of the clouds.
{\bf (d)} When averaged over 100 independent data sets, the class separability (measured by the negative General Discrimination Value GDV) depends in a monotonous way on the separation quantity S. Similarly, the degree of inter-feature correlations (measured by the RMS of the off-diagonal co-variance matrix elements) is monotonically dependent on the correlation quantity C.
\label{figure_2}}
\end{figure}

In order to investigate how the performance of different classifiers depends on the statistical properties of the data, we generate large numbers of artificial data sets with two labeled classes $c \in \left\{0,1\right\}$, in which the dimensionality $D$ of the individual data vectors $\vec{u}$, the degree of correlations $C$ between their components $u_{f=1\ldots D}$ (here also called features), and the separation $S$ between the two classes in feature space can be independently adjusted (See Figs.\ref{figure_2}(b,c) for an illustration of $C$ and $S$). To replicate some of the heterogeneity of real world data, we design our data generator as a two-level superstatistical model \cite{metzner2015superstatistical,mark2018bayesian}: The mean values $\bf \mu^{(c)}$ and covariances $\bf \Sigma^{(c)}$ of the multi-variate probability distributions $p_c(\vec{u})$ in each of the data classes $c$ are themselves random variables. They are drawn from certain meta-distributions, which are in turn controlled by the three quantities $D,S,C$ (See Methods for details, as well as Fig.\ref{figure_2}(a)).

\vspace{0.2cm}\noindent Using the General Discrimination Value (GDV), a measure designed to quantify the separability of labeled point sets (data classes) in high-dimensional spaces \cite{schilling2021quantifying}, we show that the mean separability of data classes in the DSC-model is indeed monotonously increasing with the control quantity $S$ (Orange line in Fig.\ref{figure_2}(d)), whereas the separability of individual data sets is fluctuating heavily around this mean value (Grey dots in Fig.\ref{figure_2}(d)).

\vspace{0.2cm}\noindent Moreover, we quantify the degree of correlation between the $D$ features of the data vectors in each class $c$ by the root-mean-square average of the upper triangular matrix elements in the covariance matrix $\bf \Sigma^{(c)}$. We show that this RMS-average is an almost linear function of $C$ (Blue line in Fig.\ref{figure_2}(d)) and can be varied between zero (Corresponding to independently fluctuating features, or statistical independence) and one (Corresponding to identically fluctuating features, or perfect correlations).

\subsection*{Part 3: Comparing classifiers}

\begin{figure}[ht!]
\centering
\includegraphics[width=13cm]{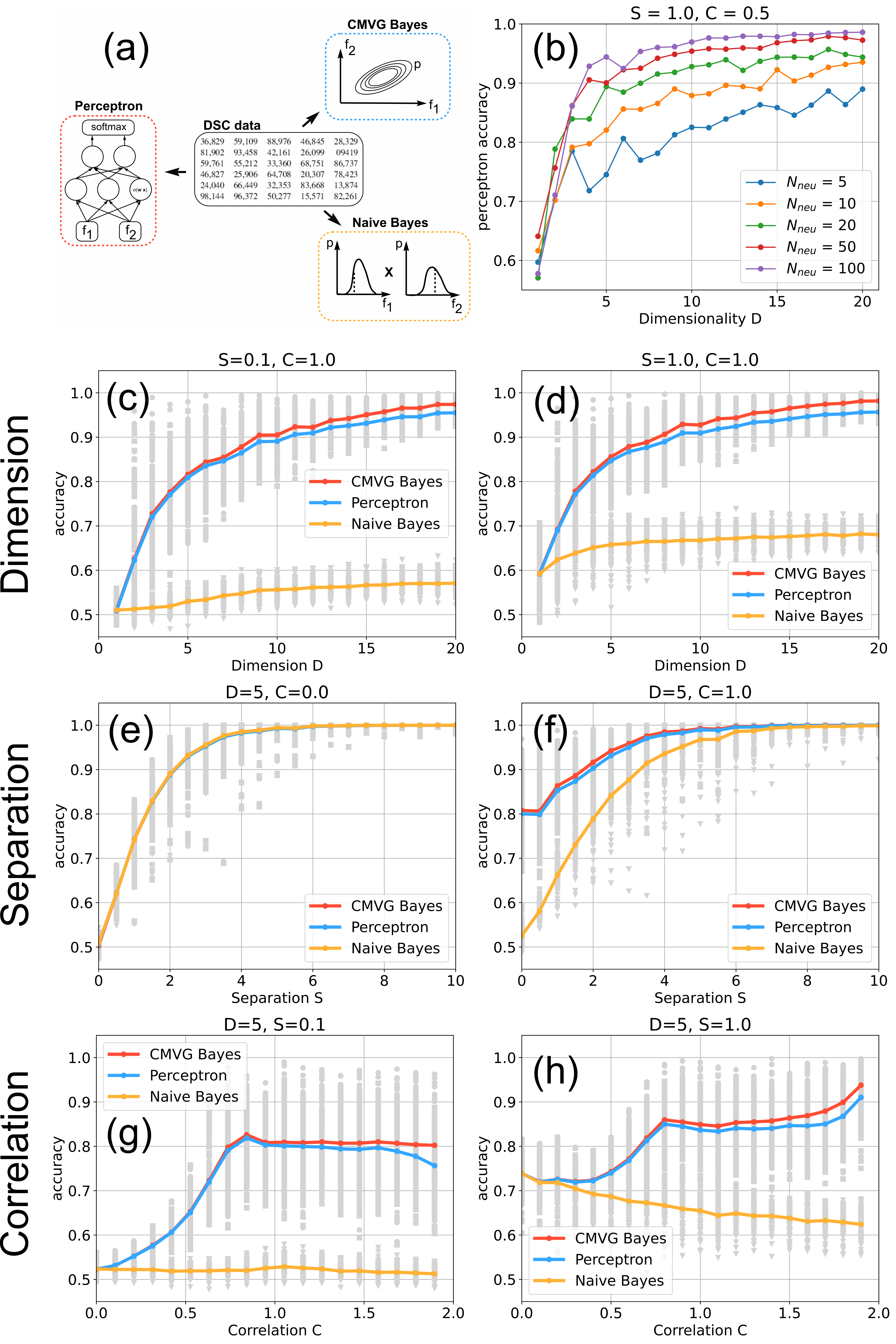}
\caption{Performance of three classifier types as a function of data statistics.
{\bf (a)} A Perceptron, a Naive Bayesian classifier and a correlated multi-variate Gaussian (CMVG) Bayesian classifier are applied to the same artificial data, controlled by the quantities $D$, $S$ and $C$. 
{\bf (b)} Accuracy of a two-layer perceptron (with 2 neurons in the second layer) as a function of the number $N_{neu}$ of neurons in the first layer, for fixed values of the separation $S=1.0$ and Correlation $C=0.5$. As the perceptron is reaching the theoretical limit of accuracy for $N_{neu}=100$, this layer size is used in the following.
{\bf (c,d)} Classifier accuracies as a function of dimension $D$ (number of features). All classifiers profit from more available features, however Perceptron and CMVG Bayes can make use of correlations and thus outperform Naive Bayes for $C=1.0$, independent from class separation $S$. 
{\bf (e,f)}  Classifier accuracies as a function of the separation $S$ between the data classes in feature space. In the case without correlations $C=0$, all classifiers reach the theoretical performance limit and thus produce identical accuracies.
{\bf (g,h)}  Classifier accuracies as a function of the correlation $C$ between features. The accuracy of Naive Bayes is degrading with increasing correlations. By contrast, Perceptron and CMVG Bayes initially profit from correlations, but abruptly reach a plateau at $C\approx 0.8$. Beyond that transition point, accuracy can further improve only for large data separation.
\label{figure_3}}
\end{figure}

Next, we apply the three classifier types to artificial data, with statistical properties controlled by the quantities $D$, $S$ and $C$. We first investigate the {\bf accuracy of the classifiers as a function of data dimensionality $D$}  (Fig.\ref{figure_3}(c,d)), considering correlated data ($C=1.0$). 

\vspace{0.2cm}\noindent When the separation of the data classes in feature space is small ($S=0.1$, panel (c)), the classification accuracy for one-dimensional data ($D=1$) is very close to the minimum possible value of 0.5 (corresponding to a purely random assignment of the two class labels) in all three models. As data dimensionality $D$ increases, all three models monotonically increase their average accuracies (colored lines), whereas the accuracies of individual cases show a large fluctuation (gray dots). However, the Naive Bayes classifier (orange line) does not perform well even for large data dimensionality, because the point clouds corresponding to the two classes are strongly overlapping in feature space. By contrast, the CMVG Bayes classifier (red line) and the Perceptron (blue line) eventually achieve a very good performance, because they can exploit the correlations in the data. The similarity of the latter two accuracy-versus-$D$ plots is remarkable, considering that these two classifiers work in completely different ways (the Bayesian model performing theory-based mathematical operations with estimated probability distributions, the neural network computing quite arbitrary non-linear transformations of weighted sums). We therefore conclude that the latter two models approach the theoretical optimum of accuracy for each combination of the control quantities $D,S,C$. 

\vspace{0.2cm}\noindent As the separation of the data classes in feature space gets larger ($S=1.0$, panel (d)), the accuracy-versus-$D$ plots are qualitatively similar to panel (c), but for one-dimensional data ($D=1$) the common accuracy is now slightly above the random baseline, at 0.6. By comparing panels (c) and (d) we note that Naive Bayes is profiting from the larger class separation, but the other two classifiers reach the theoretical performance maximum even without this extra separation.

\vspace{0.2cm}\noindent Next, we investigate the {\bf accuracy of the classifiers as a function of class separation $S$}  (Fig.\ref{figure_3}(e,f)), considering five-dimensional data ($D=5$). Without correlations ($C=0$, panel (e)), all three models show exactly the same monotonous increase of accuracy with separation $S$, starting at the random baseline of 0.5 and finally approaching perfect accuracy of 1.0. 

\vspace{0.2cm}\noindent With feature correlations present ($C=1.0$, panel (f)), the Naive Bayes classifier shows the same behavior as in panel (e), whereas the other two correlation-sensitive models now already start with a respectable accuracy of 0.8 at zero class separation.

\vspace{0.2cm}\noindent Finally, we investigate the {\bf accuracy of the classifiers as a function of the feature correlations $C$}  (Fig.\ref{figure_3}(g,h)), considering again five-dimensional data ($D=5$). For strongly overlapping data classes ($S=0.1$, panel (g)), Naive Bayes cannot exceed an accuracy of about 0.55, whereas the two correlation-sensitive models show a super-linear increase of accuracy with increasing feature correlations. However, this decrease is ending rather abruptly at about $C\approx 0.7$. Above this transition point, both models stay at a plateau accuracy of about 0.8, independent of the correlation quantity. Note that this discontinuity of the slope of the accuracy-versus-$D$ plots is likely not an artifact of the DSC data, since the RMS-average of empirical correlations versus $C$ (Fig.\ref{figure_3}(d)) did not show such an effect at $C\approx 0.7$. Moreover, the fact that functionally distinct classifiers such as CMVG Bayes and Perceptron produce an almost identical behaviour here suggests that the accuracy plateau in the strong correlation regime indeed reflects the theoretical performance maximum.

\vspace{0.2cm}\noindent As the class separation is increased ($S=1.0$, panel (h)), all three models start at a larger accuracy of about 0.75 in the uncorrelated case. Now the performance of  Naive Bayes is even declining with increasing $C$, because this model wrongly assumes uncorrelated data. The other two models show again the super-linear increase up to $C\approx 0.7$. However, now a further improvement of performance is possible with increasing correlations.

\subsection*{Part 4: Feature transformations}

\begin{figure}[ht!]
\centering
\includegraphics[width=18cm]{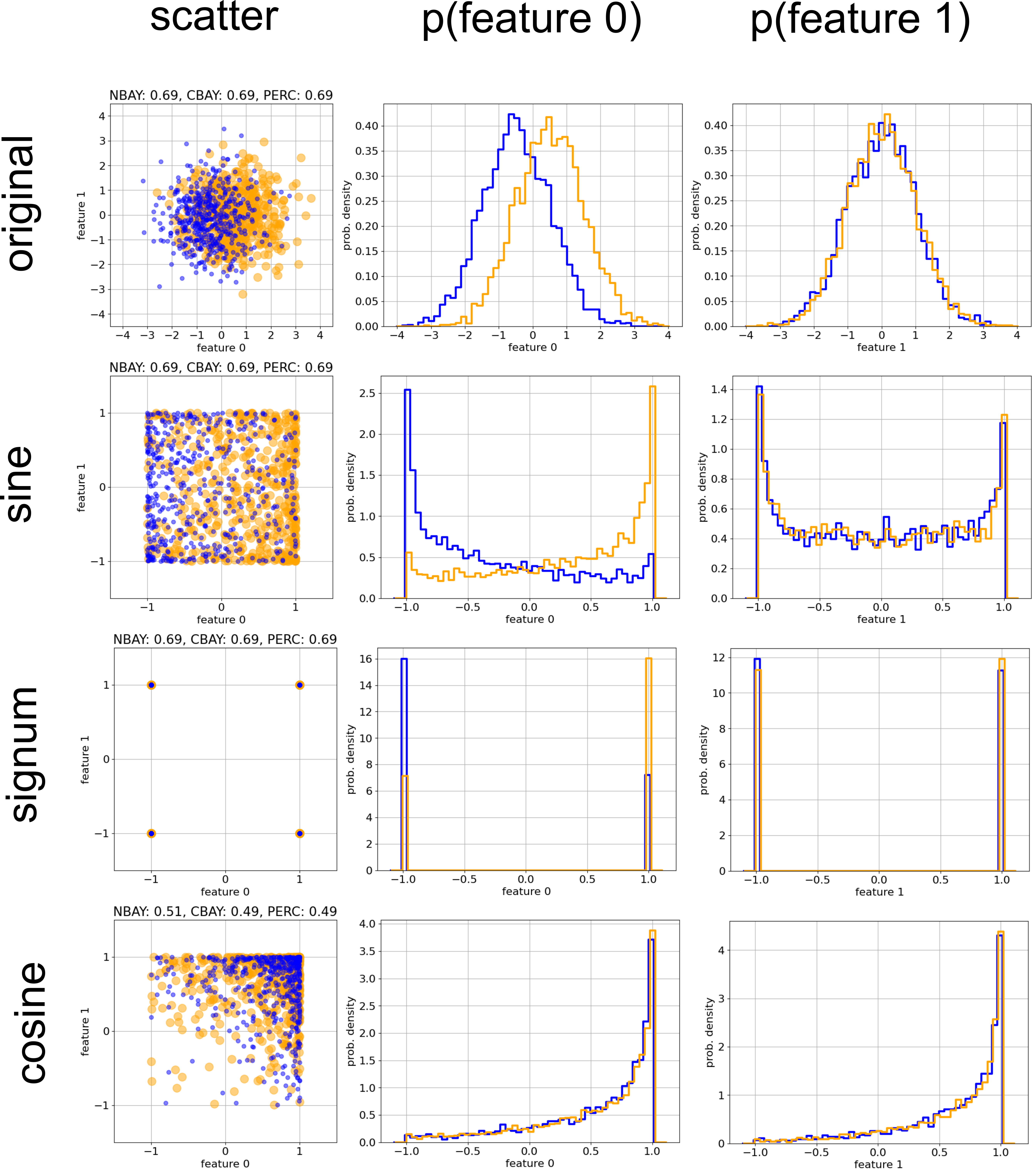}
\caption{Effect of non-linear feature transformations on the maximal classification accuracy of the Naive Bayesian model (NBAY), the CMVG Bayesian model (CBAY) and the Perceptron (PERC). We consider two partly overlapping classes (blue and orange colors) in a two-dimensional features space. Shown are a scatter plot of the data (first column), as well as the marginal probability densities of the two features (second and third column). In the case of the orginal data (first row), all three classifier types reach the theoretical accuracy limit of $\approx\!0.69$. After applying a sine-function individually to each feature (second row), the shape of the distributions changes drastically, but the accuracies remain unchanged at the theoretical limit. This remains even true after applying a signum transformation (third row), which reduces the originally continuous data to only four discrete points. However, applying a cosine transform (fourth row) reduces the accuracy to the random baseline of $\approx\!0.5$, because the two data classes now overlap completely.
\label{figure_4}}
\end{figure}

The accuracy limit is determined by the overlap of data classes, that is, by the possibility that different classes $i \neq j$ produce exactly the same data vector $\vec{x}^{\ast}$. Transformations $\vec{x} \rightarrow \vec{f}(\vec{x})$ of the input features can drastically change the distributions of data points (As an example, compare the rows in Fig.\ref{figure_4}). However, they cannot be expected to reduce the fundamental amount of class overlap, because transformations are just redirecting the common points $\vec{x}^{\ast}$ to new locations in feature space. In particular, invertible transformations can be viewed as variable substitutions in the integral Eq.\ref{cij} for the confusion matrix. They do not affect the resulting matrix values and thus leave the accuracy invariant.

\vspace{0.2cm}\noindent In order to test this expectation, we start with two overlapping Gaussian data classes in a two-dimensional feature space (Fig.\ref{figure_4}, top row), resulting in an accuracy limit of $\approx 0.69$. All three classifiers actually reach this limit with the original data as input.

\vspace{0.2cm}\noindent Next we perform simple non-linear transformations on the input data, by replacing each of the two features $x_1$ and $x_2$ with a function of themselves (in particular: $\sin$, $\mbox{sgn}$, and $\cos$). We find that the application of the $\sin$-transformation (second row in Fig.\ref{figure_4}) has indeed no effect on the accuracy of the three classifiers, even though the joint (first column) and marginal distributions (second and third column) are now strongly distorted. Even the application of the $\mbox{sgn}$-transformation (third row), which collapses all data onto just 4 possible points in feature space, leaves the accuracies invariant. This works because the two classes in our simple example can be distinguished by the sign of the $x_1$-feature, and both the $\sin$- as well as the $\mbox{sgn}$-transformation leave this information intact. By contrast, the application of the $\cos$-transformation destroys this crucial information, and consequently all accuracies drop to the random baseline of 0.5.

\vspace{0.2cm}\noindent The above numerical experiments illustrate that transformations of the input-data can reduce (by destroying information that is essential for class-discrimination), but never increase the theoretical accuracy limit, which is an inherent property of the data. Of course, the subsequent data transformations which are taking place in the layers of deep neural networks are still useful, because they re-shape data distributions until classes can be linearly separated in the final layer of the network.

\subsection*{Part 5: Sleep EEG data}

In our artificial data sets, all feature distributions were normally distributed. Moreover, it was possible to introduce extremely strong correlations between these features, which could then be exploited by two of the three classifier models. It is however unclear if the ability of a classifier to detect correlations is always crucial in real-world problems.

\vspace{0.2cm}\noindent We therefore turn in a next step to actually measured EEG data, recorded over-night from 68 different sleeping human subjects. In this case, our final goal is to assign to each 30-second epoch of a raw one-channel EEG signal one of the five sleep stages (Wake, REM, N1, N2, N3). 

\vspace{0.2cm}\noindent At our sample rate, a single epoch of EEG data corresponds to 7680 subsequent amplitudes. Such high-dimensional data vectors $\vec{x}$ are however not suitable as direct input for a Bayesian classifier, nor for a flat neural network with only $\approx 100$ neurons. For this reason, we first compress the raw data vectors $\vec{x}=(x_1,\ldots x_{7680})$ into suitable feature vectors $\vec{u}=(u_1,\ldots u_D)$ of strongly reduced dimensionality $D\approx 10$. Since we aim to develop a fully transparent classifier system, we use mathematically well-defined, human-interpretable features $u_f = G(\vec{x},\alpha_f)$, which depend on a freely tunable parameter $\alpha$. The dimensionality $D$ of the feature space is then determined by how many of these parameters $\alpha_{f=1\ldots D}$ are chosen.

\begin{figure}[ht!]
\centering
\includegraphics[width=15cm]{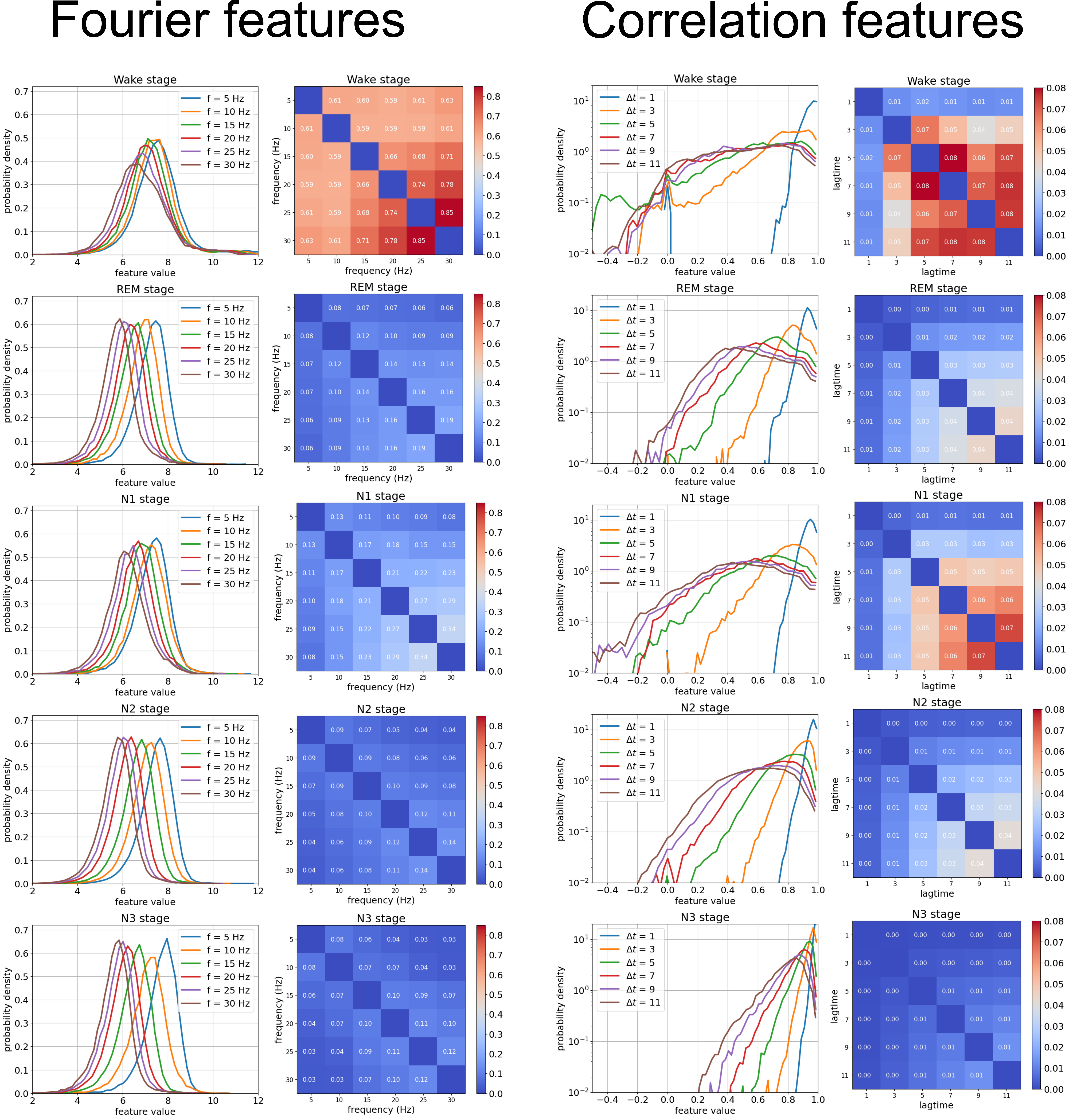}
\caption{Distributions (Columns 1 and 3)) and covariances (Columns 2 and 4) of Fourier-based and correlation-based features, extracted from EEG-recordings of humans during sleep (See methods for details). The rows correspond to the five sleep stages $s$ (Wake, REM, N1, N2 and N3). In the covariance matrices, the relatively large diagonal elements are suppressed for better visibility of the inter-feature correlations. The probability distributions $p_s(u_f)$ in row 1 are approximately Gaussian, whereas the distributions $p_s(u_{\Delta t})$ in row 3 are highly non-Gaussian. All distributions change in a systematic way with the feature parameters (frequencies $f$ of Fourier modes, lag-times $\Delta t$ of auto-correlations). Both the distributions and covariances also show characteristic differences between the sleep stages $s$, which can be  exploited for automatic classification.
\label{figure_5}}
\end{figure}

\begin{figure}[ht!]
\centering
\includegraphics[width=15cm]{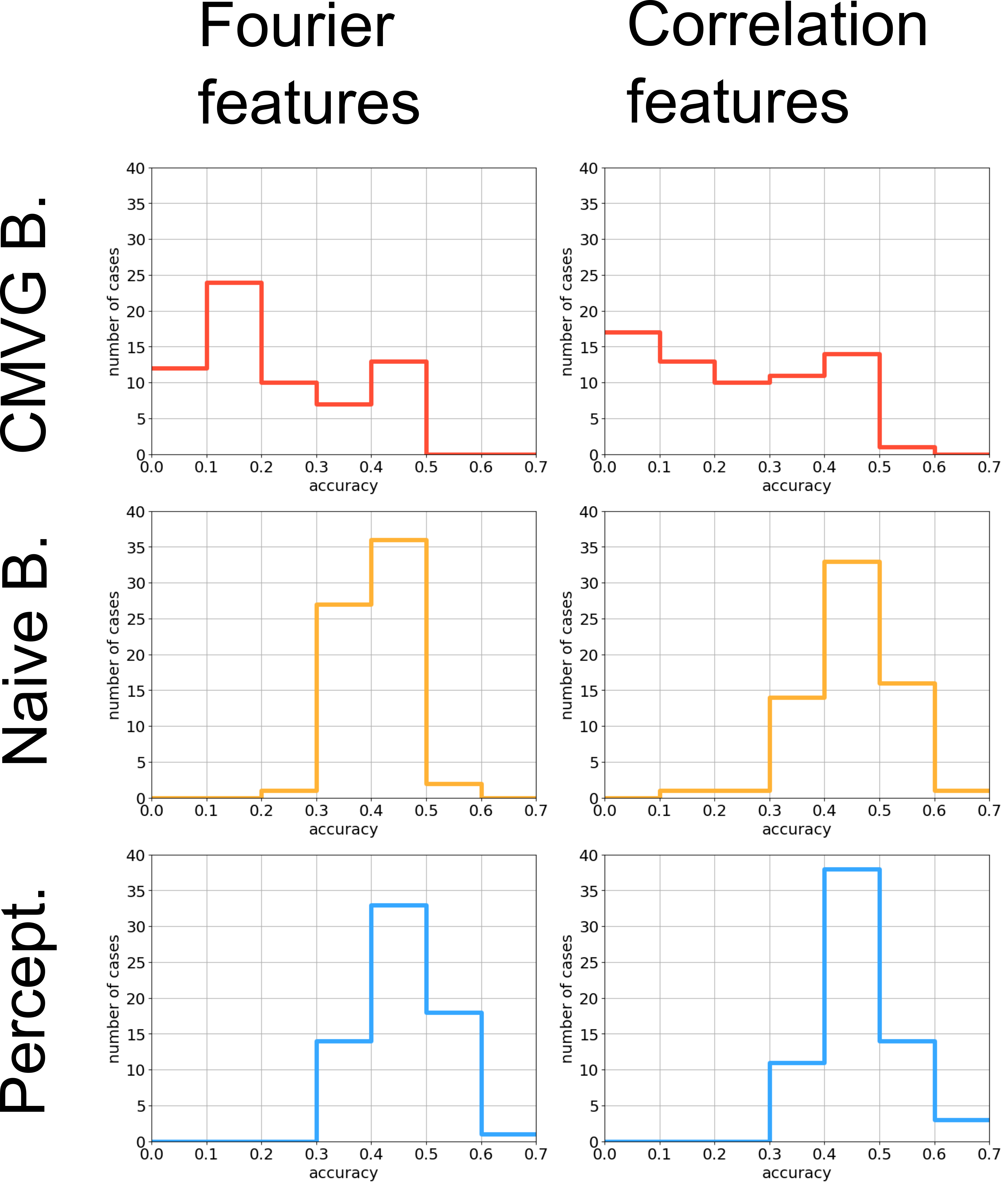}
\caption{Classification accuracies in the personalized sleep stage detection task. Full-night sleep recordings are divided into 30-second epochs, from which Fourier- and correlation features are extracted. However, instead of using the pooled data sets as in Fig.\ref{figure_5}, the three classifier models are trained and tested with data from individual subjects only. The plots show the distributions of the resulting accuracies (fraction of correctly classified 30-second epochs) over the 68 data sets. Surprisingly, the perceptron and Naive Bayes are performing about equally well, whereas CMVG Bayes fails, presumably due to the non-Gaussian feature distributions. 
\label{figure_6}}
\end{figure}

\vspace{0.2cm}\noindent The huge literature on brain waves suggests that the momentary {\bf Fourier components} of the EEG signal are suitable features for the classification of sleep stages. The parameter $\alpha$ is then naturally given by the frequency $\nu$ of the Fourier component (For details see methods). In a first experiment, we use a set of six equally spaced frequencies ($\nu_1=$5 Hz, $\nu_2=$10 Hz,$\ldots$ $\nu_6=$30 Hz). Based on training data sets that have been manually labeled by a sleep specialist, we then compute the marginal probability density functions of these Fourier features, as well as their covariance matrices, for each of the 5 sleep stages $s$ (Fig.\ref{figure_5}, left two columns). We find that within each sleep stage, the Fourier features have unimodal distributions, with peak positions and widths depending quite systematically on the frequency $\nu$. There are characteristic differences between the sleep stages (in particular the distributions are wider in the wake stage), but they are not very pronounced. In the covariance matrices, we find that the off-diagonal elements are significantly smaller than the diagonal elements (The latter have been set to zero in Fig.\ref{figure_5} to emphasize the actual inter-feature correlations), with the exception of the wake state. Also the N1 state has slightly larger inter-feature correlations compared to the REM, N2 and N3 states. 

\vspace{0.2cm}\noindent As an alternative or complement to the Fourier features, we also consider the normalized (Pearson) {\bf auto-correlation coefficients} of the raw EEG signal (Fig.\ref{figure_5}, right two columns. For details see methods). The feature parameter $\alpha$ is in this case given by the lag-time $\Delta t$, for which we choose six equally spaced values ($1,3,\ldots,11$ in units of the EEG sampling period). Since these correlation features cannot exceed the value of one by definition, the marginal distributions are highly non-Gaussian with pronounced tails towards small values. These tails show relatively strong differences between some of the sleep stages, but also surprising similarities, in particular for REM and N2. In the covariance matrices, we find the strongest inter-feature correlations in the wake and N1 stages. Again, the covariance matrices are very similar in REM and N2.

\subsection*{Part 6: Sleep stage detection}

Next, we apply our three classifier models to the above sleep EEG data. However, while the feature distributions and correlations in Fig-\ref{figure_5} were based on the global data, pooled over all 68 full-night EEG recordings, we are considering here the task of personalized sleep-stage recording. That is, the classifiers are trained and evaluated individually on each of the 68 data sets. Because the amount of training data is severely limited in this task, classification accuracies are expected to be rather low and strongly dependent on the participant. We therefore compute the distributions of accuracies over the 68 personalized data sets (histograms in Fig.\ref{figure_6}) for all three classifiers and for the two types of pre-processed features. 

\vspace{0.2cm}\noindent We find that the CMVG Bayes model is performing very poorly in this task, presumably because the feature distributions are non-Gaussian and only weakly correlated except in the wake stage. In particular, for some participants the classification accuracy is less then the random baseline of about 0.2, corresponding to consistent miss-classifications. This can happen in Bayesian classifiers when the likelihood distributions learned from the training data set do not match the actual distributions in the test data set. 

\vspace{0.2cm}\noindent By contrast, the Naive Bayes model can properly represent the non-Gaussian feature distributions by KDE approximations, and it furthermore profits from the lack of correlations. The performance of the Perceptron is comparable to that of the Naive Bayes model. Both for Fourier- and correlation-features, these two models show accuracies well above the baseline, roughly in the range from 0.3 to 0.6.

\subsection*{Part 7: Natural data clustering}

Both the ten digits in MNIST, as well as the five sleep stages in overnight EEG recordings, are human-defined classes. It is therefore unclear whether these classes can also be considered as 'natural kinds'. 

\begin{figure}[h!]
\centering
\includegraphics[width=15cm]{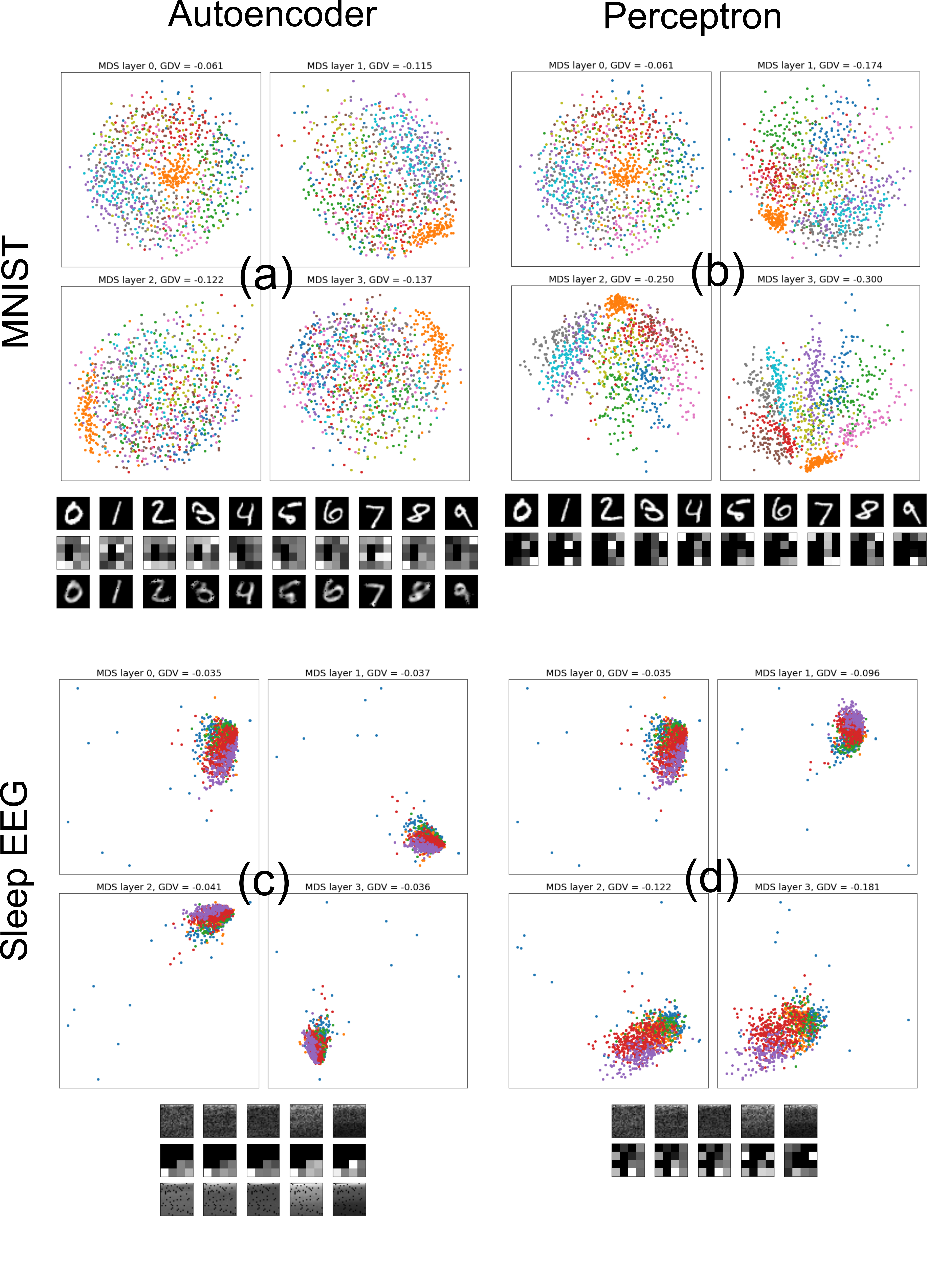}
\caption{Clustering of natural data distributions in different network layers, after supervised and unsupervised training. As example data sets we use MNIST (upper part) and the Fourier amplitudes of EEG data recorded during sleep (lower part). As simple machine learning models we use an autoencoder (left part) for unsupervised training and a perceptron (right part) for supervised training. See method section for details of models and data pre-processing.  For each combination (a,b,c,d) of model and data set, we show MDS projections of the data distributions in different layers of the network, with data classes marked by colors. The degree of class separability is quantified by the GDV in the titles of the MDS plots. Additionally, we show example input patterns (as 28x28 pixel arrays) for each data class, together with their representations in the smallest network layer 3 (as 4x4 pixel arrays). In the case of the autoencoder, we also show the reconstructed patterns.
\label{figure_7}}
\end{figure}

\vspace{0.2cm}\noindent After a suitable pre-processing that brings both data sets into the same format of 784-dimensional, normalized feature vectors (for details see Methods sections), we address this question by computing two-dimensional MDS projections, coloring the data points according to the known, human-assigned labels (In Fig.\ref{figure_7}, see the upper left scatter plot in each 2-by-2 block). Indeed, the projected data distributions show a small degree of clustering, which is also quantitatively confirmed by the corresponding GDV values (-0.061 for MNIST and -0.035 for sleep EEG data). Note that in the sleep data, a large number of extreme outliers are found which might not correspond to any of the standard classes.

\vspace{0.2cm}\noindent The purpose of classifiers is to transform and re-shape the data distribution in such a way that the final network layer (often a softmax layer with one neuron for each data class) can separate the classes easily from each other. Although, as we have shown above, these re-shaping transformations cannot reduce the natural overlap of classes (which would push the accuracy beyond the data-inherent limit), they might as a side-effect lead to a larger 'centrality' of the clusters associated with each class. This would show up quantitatively as a decrease of the General Discrimination Value (GDV) in the higher network layers of the classifier, as compared to the original input data. In order to test this hypothesis, we have trained a four-layer perceptron (see Methods section for details) in a supervised manner on both the MNIST and sleep EEG data. In the case of MNIST, we indeed observe a systematic decrease of the GDV in subsequent network layers: GDV(L0)=-0.061, GDV(L1)=-0.174, GDV(L2)=-0.250, and GDV(L3)=-0.300 (See Fig.\ref{figure_7}(b)). An analogous layer-wise decrease is found for the sleep EEG data: GDV(L0)=-0.035, GDV(L1)=-0.096, GDV(L2)=-0.122, and GDV(L3)=-0.181 (See Fig.\ref{figure_7}(d)).

\vspace{0.2cm}\noindent We finally address the question whether a natural clustering in novel, unlabeled data sets can be automatically detected, and possibly enhanced, in an unsupervised manner. For this purpose, we consider an autoencoder that performs a layer-wise dimensionality reduction of the data, and then re-expands these low-dimensional embeddings back to the original number of dimensions. During this process of 'compression' and 're-expansion', fine details of the data have to be discarded, and it appears reasonable that this might go hand in hand with a 'sharpening' of the clusters. Again, in our test case where the labels of the data points are actually known, this enhancement of cluster centrality can be quantitatively measured by the GDV. For comparability, we have used an autoencoder that has the same design as the perceptron for the first four network layers. In the case of MNIST, we indeed find that the unsupervised compression enhances cluster centrality: GDV(L0)=-0.061, GDV(L1)=-0.115, GDV(L2)=-0.122, and GDV(L3)=-0.137 (See Fig.\ref{figure_7}(a)). The behavior is similar with the sleep EEG data, except for the last layer: GDV(L0)=-0.035, GDV(L1)=-0.037, GDV(L2)=-0.041, and GDV(L3)=-0.036 (See Fig.\ref{figure_7}(c)).

\vspace{0.2cm}\noindent 

% ******************************************
% DISCUSSION
% ******************************************

\clearpage
\section{Discussion and Outlook}

In this work, we have addressed various aspects of data ambiguity: the fact that multi-dimensional data spaces usually contain vectors that cannot be unequivocally assigned to any particular class. 
The probability of encountering such ambiguous vectors is easily underestimated in machine learning, because the data sets used to train classifiers - rather than being sampled randomly from the entire space of possible data - typically represent just a tiny, pre-selected subset of 'reasonable' examples. For instance, the space of monochrome images with full HD resolution and 256 gray values contains $256^{1920 \times 1080} \approx 10^{4993726}$ possible vectors. The fraction of these images that resemble any human-recognizable objects is virtually zero, whereas the largest part would be described as noise by human observers. One may argue that these 'structure-less' images should not play any role in real-world applications. However, it is conceivable that sensors in autonomous intelligent systems, such as self-driving cars, can produce untypical data under severe environmental conditions, such as snow storms. How to deal with data ambiguity is therefore a practically relevant problem. Moreover, as we have tried to illustrate in this paper, data ambiguity has interesting consequences from a theoretical point of view.

\vspace{0.2cm}\noindent In part one, we have derived the theoretical limit $A_{max}$ of accuracy that can be achieved by a perfect classifier, given a data set with partially overlapping classes. By generating artificial data classes with Gaussian probability distributions in a two-dimensional feature space and with a controllable distance $d$ between the maxima, we verified that different types of classifiers (The CMVG Bayesian model with multi-variate Gaussian likelihoods and a perceptron) exactly follow the predicted accuracy limit $A_{max}(d)$ (Fig.\ref{figure_1}(g)). The naive Bayesian model, which cannot exploit correlations to distinguish between data classes, originally yields sub-optimal accuracies for small distances $d$, but this problem can be fixed by applying a random dimensionality expansion to the data as a trivial pre-processing step \cite{yang2021neural}. We have restricted ourselves to only two features (dimensions) for this test, because predicting the accuracy limit involves the exact computation of the confusion matrix, which in turn is an integral over the entire data space. Note, however, that for high-dimensional data with known class-dependent generation densities $p_{gen}(\vec{x}\;|\;i)$, the integral could be approximated by Monte Carlo sampling. In this case, the element $C_{ji}$ of the confusion matrix would be computed by drawing random vectors $\vec{x}$ from class $i$. The class indicator function $\hat{q}_{cla}(\;k\;|\;\vec{x}\;)$ of the perfect classifier, which is fully determined by the generation densities, yields the corresponding predicted classes $k$ for these data vectors. The matrix element $C_{ji}$ is then given by the fraction of cases where $k=j$. 

\vspace{0.2cm}\noindent In part two, we have constructed a two-level model to generate artificial test data (Fig.\ref{figure_2}). The model has high-level parameters $D$, $S$ and $C$ which control the number of dimensions (features), the average separation of the two classes in feature space, as well as the average correlation between the features. For each triple of high-level parameters $D,S,C$, a large number of low-level parameters $\mu, \Sigma$ are randomly drawn according to specified distributions, which are in turn used to generate the final test data sets. The super-statistical nature of the model allows us to prescribe the essential statistical features of dimensionality, separation and correlation, while at the same time ensuring a large variability of the test data. By using the General Discrimination Value (GDV), a quantitative measure of class separability (centrality), we have confirmed that the high-level parameter $S$ controls the class separability as intended. Moreover, the proper action of parameter $C$ was confirmed by computing the root-mean-square average over the elements of the data's covariance matrix.

\vspace{0.2cm}\noindent In part three, we have applied our three types of classifiers to the test data generated with the DSC-model. Without intra-class feature correlations ($C=0$), we find that all three models show with growing separation parameter $S$ exactly the same monotonically increasing average accuracy (Fig.\ref{figure_3}(e)). Although the exact computation of $A_{max}$ is not possible in this five-dimensional data space, the perfect agreement of the three different classifiers indicates that they all have reached the accuracy limit. When intra-class feature correlations are present ($C\neq 0$), we find by systematically varying the parameters $D$, $S$ and $C$ that the resulting accuracies of the CMVG-Bayes classifier and of the perceptron are extremely similar in all considered cases, indicating again that they have reached the theoretical accuracy limit. As expected, the naive Bayesian classifier shows sub-optimal accuracies in all cases where feature correlations are required to distinguish between the classes. In general, this analysis shows that the accuracy of classification can be systematically enhanced by providing more features (larger data dimensionality $D$) as input. Extra features that do not provide additional useful information are 'automatically ignored' by the classifiers and never reduce the achievable accuracy. Moreover, accuracy can be enhanced by providing features that are correlated with each other (larger parameter $C$), but differently in each data class. Such class-specific feature correlations can be exploited for discrimination by models such as CMVG Bayes and the perceptron, but not by the naive Bayes model. Moreover, we find that the theoretical accuracy maximum as a function of the correlation parameter $C$ shows an interesting abrupt change of slope at around $C\approx 0.8$ (Fig.\ref{figure_3}(g,h)). The origin of this effect is at present unclear, but will be explored in follow-up studies.

\vspace{0.2cm}\noindent In part four, we have investigated the effect of non-linear feature transformations, applied as a pre-processing step, on classification accuracy (Fig.\ref{figure_4}). Since the achievable accuracy in a classification task is limited by the degree of overlap between the data classes, feature transformations can certainly reduce the accuracy to below the limit $A_{max}$ (when they destroy information that is essential for discrimination), but they can never push the accuracy to above $A_{max}$. This is indeed confirmed in a simple test case where all three classifier types perform at the accuracy maximum with the non-transformed data: Applying a feature-wise sine-transformation drastically changes the data distributions $p_{gen}(\vec{x}\;|\;i)$, but leaves the accuracies unchanged at $A_{max}$. The accuracy remains invariant even under a signum-transformation, although this non-invertible operation reduces the data distributions to only four possible points in feature space. In this extreme case, most of the detailed information about the input data vectors is lost, but the part that is essential for class discrimination, namely the sign of the feature $x_1$, is retained. This example demonstrates that classification is a type of lossy data processing where irrelevant information can be safely discarded. For this reason, neural-network based classifiers usually project the input data vectors into spaces of ever smaller dimensions, up to the final discrimination layer which needs only as many neural units as there are data classes. In this context, it is interesting that biological organisms with nervous systems, relying on an efficient classification of objects in their environment for survival, have probably evolved sensory organs and filters that only transmit the small class-discriminating part of the available information to the higher stages of the neural processing chain. As a consequence, our human perception is almost certainly not a veridical representation of the world \cite{mark2010natural,hoffman2014objects,   hoffman2018interface}.

\vspace{0.2cm}\noindent In part five, we have analyzed full-night EEG recordings of sleeping humans, divided into epochs of 30 seconds that have been labeled by a specialist according to the five sleep stages. Such recordings can been used as training data for automatic sleep stage classifiers - an application of machine learning that could in the future remove a large work load from clinical sleep laboratories. In our context of data ambiguity, sleep EEG is an interesting case because different human specialists agree about individual sleep-label assignments only in 70\% - 80\% of the cases, even if multiple EEG channels and other bio-signals (such as electro-oculograms or electro-myograms) are provided \cite{fiorillo2019automated}. This low inter-rater reliability suggests that a considerable fraction of the 30-second epochs is actually ambiguous with respect to sleep stage classification, in particular when only the time-dependent signal of a single EEG channel is available as input-data. Our first goal is a suitable dimensionality reduction of the raw data, which (at a sample rate of 256 Hz) consist of 7680 subsequent EEG values in each epoch. As a pre-processing step, we map each 7680-dimensional raw data vector onto an only 6-dimensional feature vector, so that our Bayesian classifiers (Naive and CMVG) can be efficiently used. We consider as features the real-valued Fourier amplitudes at different frequencies, as well as the auto-correlation coefficients at different lag-times (Fig.\ref{figure_5}). The Fourier features are expected to be particularly useful, as it is well-known that the activity in different EEG frequency bands varies in characteristic ways over the five sleep stages. The correlation features have been successfully applied for Bayesian classification in a former study \cite{metzner2021sleep}. In our present study, we are using either Fourier or correlation features, but no combinations of those. By performing a statistical analysis of the features, we find that within the same sleep stage, the six features have significantly different marginal probability distributions. However, these distributions are quite similar in all sleep stages, so that their value for the classification task is limited. Moreover, the correlations between features, which could be exploited by the CMVG Bayes classifier and by the perceptron, turn out to be very weak, except for the Fourier features in the wake stage. Another problem is the strongly non-Gaussian shape of the marginal probability distributions in the case of the correlation features, which cannot be properly represented by the CMVG Bayes model.    

\vspace{0.2cm}\noindent In part six, we have used our three classifier models, based on the above Fourier- and correlation features, for personalized sleep stage detection. In this very hard task, the classifiers are trained and tested, independently, on the full-night EEG data set of a single individual only. Since an individual data set contains typically less than 1000 epochs (each corresponding to one feature vector), random deviations from the 'typical' sleeping patterns are likely to be picked up during the training phase. We consequently find that the accuracies vary widely between the individual data sets. As expected, the CMVG Bayes model performs badly in this task, because there are almost no inter-feature correlations present that could be exploited for sleep stage discrimination, and because feature distributions are non-Gaussian. Interestingly, both the Naive Bayesian classifier and the perceptron achieve relatively good accuracies, mainly in the range from 0.3 to 0.6. However, these accuracies may be further increased by using more sophisticated neural network architectures \cite{stephansen2018neural, krauss2021analysis}, and hence do not represent the accuracy limit.

\vspace{0.2cm}\noindent In the final part seven, we have started to explore whether the distinct classes in typical real-world data sets are defined arbitrarily (and therefore can only be detected after supervised learning), or if the differences between these classes are so prominent that even unsupervised machine learning methods can recognize them as distinct clusters in feature space. Besides the (pooled) sleep EEG data, we have used the MNIST data set to test for any inherent clustering structure. For this investigation, the individual data points, corresponding to respectively one epoch of EEG signal or one handwritten digit, have been brought into the same format of 784-dimensional, normalized vectors. Computing directly the General Discrimination Value (GDV) of the MNIST data, based on the known labels, has indeed revealed a small amount of 'natural clustering', even in this raw data distribution. This quantitative result was qualitatively confirmed by a two-dimensional visualization using multi-dimensional scaling (MDS), however the cluster structure would hardly be visible without the class-specific coloring (left upper scatter plots in Fig.\ref{figure_7}(a,b)). By contrast, no natural clustering was found for the raw sleep EEG data when the 7680 values in each epoch were simply down-sampled in the time-domain to 784 values  (data not shown). This presumably fails because the relevant class-specific signatures appear randomly at different temporal positions within each epoch, and so the Euclidean distance between two data vectors is not a good measure of their dissimilarity. However, when we instead used as data vectors the magnitudes of the 784 Fourier amplitudes with lowest frequencies, a weak natural clustering was found also in the sleep data (left upper scatter plots in Fig.\ref{figure_7}(c,d)). We have furthermore demonstrated that the degree of clustering (for both data sets) is systematically increasing in the higher layers of a perceptron that has been trained to discriminate the classes in a supervised manner (Fig.\ref{figure_7}, right column). Finally, we have used a multi-layer autoencoder to produce embeddings of the data distributions with reduced dimensionality in an unsupervised setting. It has turned out that the degree of clustering (with respect to the known data classes) tends to increase systematically with the degree of dimensional compression (Fig.\ref{figure_7}, left column). This interesting finding, previously reported in Schilling et al. \cite{schilling2021quantifying}, suggests that unsupervised dimensionality reduction could be used to automatically detect and enhance natural clustering in unlabeled data. In combination with automatic labeling methods, such as Gaussian Mixture Models, this may provide an objective way to define 'natural kinds' in arbitrary data sets.

% ******************************************
% ADDITIONAL INFO
% ******************************************

\clearpage
\section{Additional information}

\noindent{\bf Author contributions statement:}
CM has conceived of the project, implemented the methods, evaluated the data, and wrote the paper, PK co-designed the study, discussed the results and wrote the paper, AS discussed the results, MT provided access to resources and wrote the paper, HS provided access to resources and wrote the paper.
 \vspace{0.5cm} 

\noindent{\bf Funding:}
This work was funded by the Deutsche Forschungsgemeinschaft (DFG, German Research Foundation): grant SCHU\,1272/16-1 (AOBJ 675050) to HS, grant TR\,1793/2-1 (AOBJ 675049) to MT, grant SCHI\,1482/3-1 (project number 451810794) to AS, and grant KR\,5148/2-1 (project number 436456810) to PK. \vspace{0.5cm}

\noindent{\bf Competing interests statement:}
The authors declare no competing interests.  \vspace{0.5cm}

\noindent{\bf Data availability statement:}
Data and analysis programs will be made available upon reasonable request.
\vspace{0.5cm}

\noindent{\bf Ethical approval and informed consent:} The study was conducted in the Department of Otorhinolaryngology, Head Neck Surgery, of the Friedrich-Alexander University Erlangen-Nürnberg (FAU), following approval by the local Ethics Committee (323 – 16 Bc). Written informed consent was obtained from the participants before the cardiorespiratory poly-somnography (PSG).

\vspace{0.5cm}

\noindent{\bf Third party rights:}
All material used in the paper are the intellectual property of the authors.  \vspace{0.5cm}

% ******************************************
% REFERENCES
% ******************************************
\clearpage
\bibliographystyle{unsrt}
\bibliography{references}

% ******************************************
% FIGURES
% ******************************************

\clearpage

\clearpage

\clearpage

\clearpage

\clearpage

\clearpage

\clearpage

\end{document}